# A TABU SEARCH ALGORITHM WITH EFFICIENT DIVERSIFICATION STRATEGY FOR HIGH SCHOOL TIMETABLING PROBLEM

Salman Hooshmand[1], Mehdi Behshameh[2] and OmidHamidi[3]

[1]Department of Computer Engineering, Hamedan University of Technology, Hamedan Iran
s_hooshmand1@yahoo.com

[2]Department of Computer Engineering, Islamic Azad University, Toyserkan Branch, Toyserkan, Iran
mehdibehshameh@gmail.com

[3]Department of Science, Hamedan University of Technology, Hamedan, Iran
omid_hamidi@hut.ac.ir

*ABSTRACT*

*The school timetabling problem can be described as scheduling a set of lessons (combination of classes, teachers, subjects and rooms) in a weekly timetable. This paper presents a novel way to generate timetables for high schools. The algorithm has three phases. Pre-scheduling, initial phase and optimization through tabu search. In the first phase, a graph based algorithm used to create groups of lessons to be scheduled simultaneously; then an initial solution is built by a sequential greedy heuristic. Finally, the solution is optimized using tabu search algorithm based on frequency based diversification. The algorithm has been tested on a set of real problems gathered from Iranian high schools. Experiments show that the proposed algorithm can effectively build acceptable timetables.*

*KEYWORDS*

*Timetabling. Tabu Search. Diversification .Graph . Scheduling*

## 1. INTRODUCTION

Scheduling high school lessons has been considered to be one of the main concerns of schools' staff before beginning of the term. Manual timetabling is tedious and generally takes several weeks to build an acceptable timetable. The main difficulty is related to the size of the problem; the algorithm should consider several conflicting criteria and conditions often for a large number of classes, teachers and courses. Moreover, the structure of timetable as well as criteria of quality is different between countries or even within schools inside a country. For these reasons, the problem of building high-school timetables has been extensively studied in operations research community.

The solution techniques ranging from graph coloring heuristic to complex metaheuristic algorithms. Using heuristics is justified since the problem is known to be complex and difficult and exact solutions would be possible only for problems of limited sizes [1]. Examples of these algorithms include graph coloring heuristics [2][3], Tabu Search [4][5] and simulated annealing [6][7]. A different approach is constraint logic programming, as in [8][9] and sometimes a CLP framework is exploited [10]. For comprehensive survey of the automated approaches for timetabling, the reader is referred to [11][12].

The aim of this research was to develop a method to build timetables for Iranian high schools. The algorithm has been changed several times to consider characteristics of different schools. It consists of several heuristics to obtain the initial solution. Then a tabu search algorithm with effective neighborhood exploration and diversification strategy is applied to optimize the solution.

The paper is organized as follows: section two formulates the problem, section three describes the graph based algorithm to group lessons and algorithm to generate the initial solution. Then section four details components of optimizing tabu search algorithm and its components namely, neighborhood structure and diversification strategy. The next section presents our dataset and results of experiments on them. Finally section six, argues about results and future works.

## 2. The Problem Considered

The problem consists of a set of classes *C*, which represents groups of students with the same curricula and a set of teachers *T*, who have to teach set of subject *S*, as courses to classes. Each course has a weekly structure which indicates the length of the course's lessons. For example a course meets five hours a week may have structure of 2-2-1 meaning that two lessons have two hours long and the other lasts one hour. A period is a pair composed of a day and a timeslot; There are *p* periods being distributed in *d* days and *h* daily timeslots ($p=d\times h$). In addition, there are a set of curricula *CR*, where each curriculum denotes set of same subjects that must be taught to some classes. Therefore, each lesson can be represented using a set of <*t*, *cl*, *s*> tuples which have to be met at the same time. In other words a lesson can be defined by the following definition:

$$l = \{< t, cl, s > | t \in T, cl \in C, s \in S \}$$   Eq. 1

There are two kinds of lessons: *Simple lessons* and *Blocks*. In *simple lessons*, which are the common case, a single teacher teaches a single subject to a class. In contrast, in a *block* several teachers teach several courses to several classes. In other words, a *block* consists of several lessons which have to be scheduled at the same time. There are two special kinds of bocks in Iranian high schools known as *Half-Switch* and *Double-Lesson.* In *Half-Switch,* two teachers teach to two classes in first half of period and switch their classes in the second half. In *Double-Lessons* a teacher teaches a single lesson to two or more classes at the same period. Fig . 1 depicts sample of these two kinds of blocks.

| Class1 | | Class2 | |
|---|---|---|---|
| Language | Geography | Geography | Language |

(a)

| Class1 | Class2 |
|---|---|
| Chemistry ||

(b)

Fig 1: Types of blocks in Iranian schools; (a) Half-Switch, where class 1 has language course in first half and Geography in the second half of the period and class 2 is vice-versa (b) Double-Lessons, where class 1 and class 2 students has Chemistry with a single teacher at the same time.

The problem can be described as a constraint satisfaction problem, where lessons of a set of courses should be scheduled into a weekly timetable, in accordance with a given set of constraints [1].

The constraints can be categorized to hard and soft ones. A feasible timetable is one in which lectures scheduled so that the hard constraints are satisfied. In addition, a timetable satisfying hard constraints incurs a penalty cost for the violations of soft constraints which should be minimized. The constraints used in our algorithms are as follows:

$H_1$. **Conflicts**: No two lessons should occur at the same time in a class and no teacher scheduled to teach two lessons at the same time.
$H_2$. **Availability**: Teachers and classes should be scheduled only at available periods.
$C_1$.**Class Gap**: No gap should exist between lessons of a given class in a day;
$C_2$.**Teacher Gap**: It is ideal that no gap exists between lessons of a given teacher in a day.
$C_3$.**Compactness**: Sessions of a course should not meet in same or consecutive days.
$C_4$.**Balance**: There should be balanced combination of simple and hard lessons for each class in all days according to curricula.
$C_5$.**Completeness:** All lessons should be scheduled in the timetable. Table 1 summarizes a number of symbols and variable definitions used in the problem.

Table 1: Notations used in the problem

| Symbols | Description |
|---|---|
| $C$ | Set of classes, $C = \{c_1, c_2, \ldots, c_n\}$ |
| $T$ | Set of teachers, $T = \{t_1, t_2, \ldots, t_m\}$ |
| $S$ | Set of subjects, $S = \{s_1, s_2, \ldots, s_k\}$ |
| $L$ | Set of lessons, $l = \{<t, cl, s> \mid t \in T, cl \in C, s \in S\}$ |
| $d$ | The number of working days per week |
| $h$ | The number of timeslots per day |
| $p$ | The total number of periods, $p = d \times h$ |
| $CR$ | Set of the curricula, $CR = \{Cr_1, Cr_2, \ldots, Cr_f\}$ |
| $cplx_{i,k}$ | Whether lesson$_i$ is a complex one for class$_k$. if so, $cplx_{s,k} = 1$ otherwise $cplx_{s,k} = 0$ |
| $Sch_{li}$ | Whether lesson $l_i$ is scheduled; $sch_i = \begin{cases} 0 & if \ \exists <li, p> \in Q, p \neq -1 \\ 1 & otherwise \end{cases}$ |
| $same_{li,lj}$ | if lessons $l_i$ and $l_j$ belong to same course, $same_{li,lj} = 1$ otherwise $same_{li,lj} = 0$ |
| $cls_{li,c}$ | If class $c$ involves in lesson $l_i$, $cls_{li\,c} = 1$, otherwise it is 0. |
| $tch_{li,t}$ | If teacher $t$ involves in lesson $l_i$, $tch_{li\,t} = 1$, otherwise it is 0. |
| $avl_{i,p}$ | Whether teachers and classes involved in lesson $i$ are available at period $p$. if so, $avl_{i,p} = 1$ otherwise $avl_{i,p} = 0$ |
| $A_{c,d}$ | Set of lessons scheduled to held in class $c$ on day $d$ |
| $clsgp_p(c_i)$ | Whether a gap exists in class $c_i$ schedule at period $p$<br>$clsgp_p(c_i) = \begin{cases} 1 & if \ \exists \ (p1 \in p, p2 \in p), p1 < p < p2 \wedge (p1 \bmod d = p2 \bmod d = p \bmod d) \wedge \\ & (CPB_{c_i\,p1} \neq -1 \wedge CPB_{c_i\,p} = -1 \wedge CPB_{c_i\,p2} \neq -1) \\ 0 & otherwise \end{cases}$ |
| $tchgp_p(t_i)$ | Whether a gap exists in teacher $t_i$ schedule at period p $$tchgp_p(t_i) = \begin{cases} 1 & \text{if } \exists\ (p1 \in p, p2 \in p)\ , p1 < p < p2 \wedge (p1 \bmod d = p2 \bmod d = p \bmod d) \wedge \\ & (TPB_{t_i\,p1} \neq -1 \wedge TPB_{t_i\,p} = -1 \wedge TPB_{t_i\,p2} \neq -1) \\ 0 & \text{otherwise} \end{cases}$$ |
| $compact$ $(l_i)$ | Whether lesson $l_i$ schedule is compact. $$compact(l_i) = \begin{cases} 1 & \text{if } \exists\ (<li, p1> \in Q, <lj, p2> \in p)\ , p1 \neq -1 \wedge p2 \neq -1\ (|p1 \bmod d - p2 \bmod d| \leq 1) \wedge \\ & same_{li\,lj} = 1) \\ 0 & \text{otherwise} \end{cases}$$ |
| $Un$-$balanced_d(c_i)$ | Whether a lesson scheduled for class $c_i$ at day d are too complex and not balanced. $$unbalanced_d(c_i) = \begin{cases} 1 & \sum_{l=1}^{|L|} cplx_{l,d} \times A_{l,d} > \left\lceil \frac{h}{2} \right\rceil \\ 0 & \text{otherwise} \end{cases}$$ |

## 2.1. Solution Representation

Many researchers have adopted direct representation of timetable as teachers' timetable[13] or classes' timetable [1]. Although these representations are simple to implement, they are not intrinsically suitable for complex structure of lesson blocks in our problem. Thus we adopted a representation which can conveniently model various kinds of blocks. Having previous definitions, the solution to the problem can be presented as a set of $<l, p>$ pairs where $l$ is a lesson and $p$ represents period of the week when lesson meets.

$$Q = \{<l, p> | l \in L, p \in P \cup \{-1\}\} \qquad \text{Eq. 2}$$

In above definition, $l$ is set of all lessons and $P$ is the set of all periods in each week. If a lesson is currently unscheduled, -1 is assigned as its period.

### 2.1.1 Redundant timetable representations

To improve performance of algorithm, the timetable is represented in two other data structures in parallel:

- $TPB_{t \times p}$: where $t = |T|$, $p = |P|$ and $TPB_{i,j}$ denotes the lesson which teacher $i$ teaches at period $j$. If the teacher is unscheduled at that period the value is -1
- $CPB_{c \times p}$: where $c = |C|$, $p = |P|$ and $CPB_{i,j}$ refers to the lesson which meets at period $j$ in class $i$. If the class is unscheduled at that period the value is -1

It is notable that the value of TBP is valid only if we have no teacher conflict, no teacher exists scheduled to teach two lessons at the same time. Similarly CPB is valid only if at most one lesson scheduled to be met in a class at each time. These conditions are always met during our algorithm. Nevertheless, Eq. 1 can model every possible timetable regardless of these infeasibilities.

## 2.2 Objective Function

It is obvious that meeting all constraints are ideal. In real world, except for the first and second constraints which are considered as hard, all other criteria are somehow violated. In a feasible solution, following constrains are always satisfied:

- $\nexists\ (<l1, p> \in Q, <l2, p> \in Q, c \in C\ , cls_{l1\,c} = 1 \wedge cls_{l2\,c} = 1) \wedge$
  $\nexists\ (<l1, p> \in Q, <l2, p> \in Q, t \in T\ , tch_{l1\,t} = 1 \wedge tch_{l2\,t} = 1)$
- $\forall <l, p> \in Q, avl_{l\,p} = 1$

To evaluate the current solution according to all soft constraints, $C_1$-$C_5$, following objectives can be defined as:

- $F_1(Q) = \sum_{c=1}^{|C|} \sum_{p=1}^{|P|} clsgp_p(c)$
- $F_2(Q) = \sum_{t=1}^{|T|} \sum_{p=1}^{|P|} tchgp_p(t)$
- $F_3(Q) = \sum_{l=1}^{|L|} compact(l)$
- $F_4(Q) = \sum_{c=1}^{|C|} \sum_{d=1}^{D} unbalance_d(c)$
- $F_5(Q) = \sum_{li=1}^{|L|} sch(l_i)$

Where, $F_i$ counts total number of violations of constraint $C_i$.

Having defined above formula, we can then calculate the cost for a given candidate feasible solution $Q$ according to the objective function F defined in Eq. 3.

$$F(Q) = \sum_{i=1}^{5} w_i \times F_i(Q)$$   Eq. 3

Where $w_i$ is the weight associated with each objective and the goal is then to find a feasible solution $Q^*$ such that f($Q^*$)⩽f($Q$) for all Q in the feasible search space.

## 3. Solution method

Our algorithm for building a timetable has three phases. Phase 0 (Section 3.1) which groups special sessions to form blocks which are easier to schedule; Phase 1 (Section 3.2), which constructs a feasible timetable using a fast greedy heuristic and Phase 2 (Section 3.3), which improves the initial timetable using tabu search algorithm [14] with diversification strategy. The diversification strategy is based on frequency measures of movements in search space.

### 3.1 Pre-Scheduling

As mentioned in section 2, existence of blocks is one of Iranian schools' characteristics. Due to rather complex nature of them, automatic detection of them is desirable. Fig.2. depicts some other types of blocks which algorithm tries to find in input data, namely half-Loop andhalf-chain; In half-loop a set of n teachers have lessons with n classes while classes as well as teachers are fully busy at block's assigned period. In half-chain, some teachers (Fig 2.b.) or classes (Fig 2.c) are not fully scheduled at that period.

| Class 1 | | Class 2 | | Class 3 | |
|---|---|---|---|---|---|
| Language | Geography | Geography | Art | Art | Language |
| (a) | | | | | |

| Class 1 | | Class 2 | | Class 3 | |
|---|---|---|---|---|---|
| Language | Geography | Geography | Art | Art | Math |
| (b) | | | | | |

| Class 1 | | Class 2 | | Class 3 | |
|---|---|---|---|---|---|
| - | Geography | Geography | Art | Art | - |
| (c) | | | | | |

Fig2: Other types of blocks in Iranian schools; (a) Half-Loop, (b) Half-Chain (C) special kind of Half-Chain

The proposed algorithm (Fig 3) uses a graph search for this task. The buildGraph procedure gets set of sessions with length one (one hours long) and creates the graph which will be traversed to find patterns of above mentioned blocks. Each node of this graph consists of two or one lessons. Sessions $L_1$ and $L_2$ are combined if they are taught to same class and their teachers have at least one available period in common. If some sessions cannot be combined they form nodes with just one session hopefully to form Half-chain blocks (Fig 2.c). An edge exits between two nodes if they have a teacher in common. Having defined the graph, we look for patterns of blocks. Each block can be presented as a path in graph. These paths are stored in set P. First, Find_HalfSwitch subroutine searches for most important type of blocks ,Half-Switch, in the graph. These are two adjacent nodes in graph with two teachers in common. The paths related to this blocks are stores in P and Find_Halfloopprocedure tries to find loops in the graph. A depth first search method used to detect loops in G. The loop is admissible if no node of it appeared previously in P. Finally, Find_HalfChaintraverses the graph to find half-chains; To do so, it starts from nodes with just one associated session and visits nodes until reaching another node with same structure. The method checks feasibility of this path. For example suppose that we have a path with nodes { <$class_1$-art, $class_1$-math>, <$class_2$-Math, $class_2$-Music>, <$class_3$-Geography, $class_3$-Math>} this path is not feasible since math teacher should teach to $class_2$ and $class_3$ simultaneously. It is notable that Half-Chains should be marked to be scheduled preferably at first or last session of a day, since putting them in the middle of the day causes gap for teachers and/or classes' timetable.

### 3.2. Initial solution

We have adopted a sequential greedy heuristic to build the initial solution; starting from an empty table the algorithm repeatedly schedules the lesson with most priority at its most suitable period. Priority of a lesson is defined as the inverse of number of possible period for it. Lesson *l* can be scheduled at period *p* if it is available at it,i.e. $avl_{l,p} = 1$, while assigning *p* to *l* does not violate any of constraints. Then we have to choose a period from candidate periods. For each period we count number of unscheduled lessons who can be scheduled at that period later. The period with smallest value of this number will be chosen. This initial timetable is used as the starting point for optimization phase.

| | |
|---|---|
| 1. | Procedure buildblocks(L) |
| 2. |     G ←buildGraph(V,E, L); |
| 3. |     P ← {} |
| 4. |     P ← P ∪Find_HalfSwitch(G); |
| 5. |     P ← P ∪Find_Halfloop(G, P); |
| 6. |     P ← P ∪Find_HalfChain(G, P); |
| 7. |     Return P; |
| 8. | End buildblocks |

Fig 3: Block detection algorithm

### 3.3. Optimizing the solution

This algorithm (Fig 5)starts with initial solution obtained in previous phase and optimizes the solution by applying an algorithm that is an adaptation of general Tabu Search technique to our problem definition. Tabu search (TS) is an iterative procedure designed for the solution of optimization problems. It explicitly makes use of memory structures and responsive exploration to guide a hill-descending heuristic to continue exploration without being confused by the absence of improvement movements. Interested readers are referred to [14] for a detailed description of *TS*.

International Journal of Computer Science & Information Technology (IJCSIT) Vol 5, No 4, August 2013

### 3.3.1. Search space (X) and objective function

The search space of our algorithm is the set of schedules satisfying all hard constraints. For each solution *s* we have a set of scheduled lessons and a set *U(s)* of unscheduled lessons. The objective function is to minimize |*U(s)*| and other soft constraints penalties as described in the objective function (Section 2.2)

### 3.3.2. The neighborhood N(s)

A solution *s'* ∈*X* is a neighbor of solution *s* ∈*X* if it can be obtained from *s* by changing period of set of lessons in *s*. A move is composed of 3 steps and inspired by [10].

Step 1. Take a lesson *l* from set of candidate lessons, *R*, and assign it a new period *t*.
Step 2. Remove from schedule the lessons for which the assignment of period *t* to *l* violates any hard constraint.
Step 3: Consider each new element of *U(s')* for its inclusion in new period with no conflict with any scheduled lesson.

In other words move *M* can be described as a set:

$$M = \{< l, p, dir > | \; l \in L, p \in P, dir \in \{in, out\}\} \qquad \text{Eq. 4}$$

It is notable that the definition is so general that can incorporate much diverse range of moves. Here we have used two types of moves.

1: **Out-In move**: This moves one of unscheduled lessons to table. i.e. in step 1, *R* is *U(s)*.
2: **Intra move**: This type of move changes the period of a scheduled lesson in the table to a new period inside table i.e. *R* in step 1 is set of all lessons except *U(s)*.

*Out-In* moves favor minimizing unscheduled lessons ($f_5$ function) as it has the highest weight among criteria and is quite effective as will be shown in experiments (section 4). However we found that as the optimization phase proceeds the set *R* eventually becomes so small that we trap in a local minimum and need a way to escape this region. The Intra move proposed to overcome this problem by rescheduling previously scheduled lessons at a new period. Fig 4 depicts samples of moves.

| Period\Class | Class 1 | Class 2 | | Class 3 | | Class 4 | Unscheduled |
|---|---|---|---|---|---|---|---|
| P0 | English | Art | | Chemistry | | Mathematic | History |
| P1 | | Chemistry | | History | | English | |
| P2 | Physic | Language | Geography | Geography | Language | | |

(a)

| Period\Class | Class 1 | Class 2 | | Class 3 | | Class 4 | Unscheduled |
|---|---|---|---|---|---|---|---|
| P0 | English | Art | | Chemistry | | Mathematic | History |
| P1 | | Chemistry | | History | | English | |
| P2 | Physic | Language | Geography | Geography | Language | | |

(b)

Fig 4: Sample of moves (a) Out-In move (b) Intra move

### 3.3.2.1 Switch between move types

Move type is selected at the beginning of each iteration (line 9). Out-In moves are always preferred except when we have not found better solutions for a long time (IntraActivationiterations). These moves continue for successive IntraDepth iterations. The IntraDepthparameter is adaptive; its initial value is 0 and increases by one at the start of intra-move activation period (line 8). This value becomes 0 when the best solution updates (line 32). The process is cyclic and whenever a multiple of IntraActivation non-improvement iterations occurs activates again.

### 3.3.3. The tabu lists

Since each move consists of moving lessons into or out of timetable, the tabu list consists of ($l$, $t$) pairs of lessons and the corresponding periods in which they were scheduled before move. A move is considered tabu if any of lessons involving in it, came back to their previous positions. In other word we can say move $M$ is tabu if following holds:

$$\exists\ (< l, p, d > \in M\ ,\ < l, p > \ \epsilon\ Tabu\_list) \qquad \text{Eq. 5}$$

After selecting a move, *m*, its members are put in tabu list for next *tabuTenure(m)* iterations. We have used lists with a random length selected from numbers in range (*0.25l*, *2l*) where *l* is square root of number of lessons as proposed by [10].

```
1.  Procedure ImproveTimeTable(T, L, divActivation, iterationsDiv) {
2.  buildblocks(L)
3.  T* ← T; TabuList = {}; noimprovementIteration ← 0; iteration ← 0;IntraDepth ← 0;
4.  While termination criteria not reached
5.      Iteration++;
6.      Δ ← ∞
7.      If (noimprovementIteration mod IntraActivation = 0)
8.          IntraDepth++
9.      If (noimprovementIteration mod IntraActivation<IntraDepth)
10.         Movetype ← Out-In;
11.     Else
12.         Movetype ← Intra;
13.     For all movements m such that (T ⊕ m) ∈ N(T)
14.         Penalty ← 0;
15.         If (noimprovementIteration mod divActivation<iterationsDiv and
16.             Iteration >divActivation)
17.             Penalty ← computePenalty(m);
18.         End if
19.         Δ' ← f(T ⊕ m) - f(T);
20.         If ((Δ' + Penalty <Δ ) and (m ∉TabuList)) or (f(T ⊕ m) < f(T*)) then
21.             bestMove ← m;
22.             Δ = Δ';
23.             if (f(T ⊕ m) > f(T*)) then Δ = Δ + Penalty;
24.         End if
25.     End for
26.     updateLongTermMemory(bestMove, T);
27.     T ← T ⊕ bestMove;
28.     MakeTabu (bestMov);
29.     updateTabuList(bestMov, iteration);
30.     If (f(T) < f(T*)) then
31.         T* ← T; noImprovementIterations = 0;
32.         IntraDepth = 0;
33.         initializeLongTermMemory();
34.     Else
35.         noImprovementIterations++;
36.     End if
37. End while
38. Return T*
39. End ImproveTimeTable
```

Fig 5: pseudo code of optimization algorithm

### 3.3.4. Aspiration criterion

The aspiration criterion defined is that the movement will lose its tabu status if its application produces the best solution found so far.

### 3.3.5. Diversification

To encourage method to explore previously unvisited regions of search space and minimize the risk of entrenching in local optima, a diversification strategy is used. In our algorithm a transition based long term memory exploited to store the frequency of movements involving each lesson and period. This memory updates at each iteration (line 26) and clears when a better total solution is found (line 33). When diversification is active, this long term memory involved in evaluating moves and moves which differ most from previous ones are prioritized. This is done through using penalty in evaluating moves (line 17). In the following paragraphs a description of the proposed long term memories and how they are used to compute penalties in the diversification strategy is presented.

29

International Journal of Computer Science & Information Technology (IJCSIT) Vol 5, No 4, August 2013International Journal of Computer Science & Information Technology (IJCSIT) Vol 5, No 4, August 2013

**Transition-based long term memory**:

In this kind of memory, information stored in a matrix $Z_{l \times p}$ where $l = |L|$, $p = |P|$ and $Z_{i,j}$ denotes how many times lesson *i* has moved to period *j*. using these values we define move ratio. First we define:

$$\bar{z} = \max\{z_{i,j} | i \in L, j \in P\} \qquad \text{Eq. 6}$$

Accordingly move ratio of a lesson is defined as

$$\varepsilon_{i,j} = \frac{z_{i,j}}{\bar{z}} \qquad \text{Eq. 7}$$

And penalty of move *M* is calculated as below:

$$Penalty(M) = \frac{\sum_{i=1}^{n} \varepsilon_{M_i,P_i}}{n} \times f(T) \qquad \text{Eq. 8}$$

Where n = |M| and f(T) is current cost function. It can be seen that more frequent moves get more penalty. The diversification strategy is applied whenever number of iterations without improvements reaches a threshold called divActivation. Diversification strategy lasts for iterationsDiv successive iterations. The process is cyclic as explained for intra move activation strategy (section 3.3.2.1). Both divActivation and iterationsDivare input parameters.

### 3.3.6. Selection of moves

Because of performance considerations, neighborhood used in this algorithm is not full. We explore a subset of N(S) in a two-step procedure. First, candidate lessons for current move type (as explained in section 3.2.2) are sorted according to inverse of number of possible assignment periods. Lessons of this sorted set are selected in turn and the first lesson which can be scheduled at a period improving the objective function (considering the penalty if diversification is active) is selected. This move is not tabu or the aspiration criterion can be applied. If no lesson can improve the cost function, a random lesson is chosen and the best move based on that lesson is done.

## 4. Experiments

### 4.1. Problem instances and experimental protocol

Experiments were done on set of three Iranian high-schools with typical 3 or 4 periods a day and some block lessons according to curriculum.
**Error! Reference source not found.**Table2 introduces some of the characteristics of the instances as number of classes, teachers and lessons. In addition, schools' data can be compared by their sparseness ratio (sr) considering the total number of lessons (#lessons) and the total number of available periods for them (p); Where it is more difficult to find a feasible solution for problems with lower sr values [15]. The sparseness ration can be defined as:

$$sr = \frac{\#a}{\#lessons \times p}$$

3030

International Journal of Computer Science & Information Technology (IJCSIT) Vol 5, No 4, August 2013

Table2: characteristics of schools

| School | Classes | Teachers | Lessons | Blocks | Day Periods | Working Days | Sparseness Ratio (*sr*) |
|--------|---------|----------|---------|--------|-------------|--------------|-------------------------|
| TA | 12 | 43 | 171 | 11 | 3 | 6 | 0.34 |
| AL | 12 | 42 | 152 | 18 | 4 | 6 | 0.22 |
| DE | 9  | 24 | 72  | 5  | 3 | 4 | 0.32 |

The objective of experiments was to verify effectiveness of different components of the proposed algorithm on a set of real datasets. Average results of 10 independent executions (different random seeds) on each instance for different configuration of tabu search components and instances were computed (in all experiments parameters had these fixed values: *divActivation* = 20, *iterationsDiv* = 5, *IntraActivation* = 40; $w_1$ = 100, $w_2$= 40, $w_3$= 30, $w_4$= 60, $w_5$= 1000) and the algorithm stops after 3000 iterations. In the following sections the tabu search implementation with no diversification and intra-moves willbe referred as *TS*, while the implementation with diversification will be referred as *TSD*; another implementation, which incorporates intra-moves called *TSI* and finally *TSDI* refers to considering both diversification and intra moves. The results are shows in Table 3.

Table3: Results of applying different tabu search strategies on
Cost function (F(Q)) and percent of decrease on initial cost (%IMP)

| TestCase | Initial algorithm | Without Diversification | | | | With Diversification | | | |
|---|---|---|---|---|---|---|---|---|---|
| | | Without Intra moves (TS) | | With Intra moves (TSI) | | Without Intra moves (TSD) | | With Intra moves (TSDI) | |
| | | *F(Q)* | *%IMP* | *F(Q)* | *%IMP* | *F(Q)* | *%IMP* | *F(Q)* | *%IMP* |
| DE | 26 000 | 8 680  | 66% | 3 240  | 87% | 2 100 | %91 | 1 050 | 96% |
| TA | 37 000 | 12 082 | 67% | 5 662  | 84% | 3 248 | %91 | 2 650 | 93% |
| AL | 61 000 | 13 682 | 77% | 11 814 | 80% | 3 326 | %94 | 2 880 | 95% |

## 4.2 Analysis of Optimization phase

It is obvious that both using diversification as well as intra moves decrease cost function considerably. While the tabu solution without proposed tabu components optimizes the initial solution by 70% on average, intra moves and diversification, used independently decrease the cost function by 83.6% and 92% respectively. It shows that on average, the diversification strategy is more effective than intra moves. However it is interesting that combination of intra moves and diversification strategy yields 24% lower cost functions in comparison with using diversification strategy alone(from average cost of 8647 for *TSD* to 6580 for *TSDI*). This decrease can be explained by the type of moves presented in section 3.2.2; as process progresses, number of unscheduled blocks decreases and just using out-in moves causes repeated type of moves. Thus intra-moves adopted to add more diversification in the search space. These moves generally increase cost function according to high weight of $w_5$(1000). It is due to the fact that changing period of a block inside table usually causes some conflicts with other blocks which must be moved out of table (For example intra move in fig 2(b) increases unscheduled blocks by one). These exited blocks increase the size of |*U(s)*| which adds more diversity to typical out-in moves.
The algorithm's performance varies between different schools; The *TSI* algorithm (Intra-moves component only) works best on simpler schools, *DE* and *TA* schools with lower sparseness ratio. In the other hand, performance of *TSD* algorithm which incorporates just diversification and no intra-moves peaks for the hardest problem *AL* with lowest sparseness ratio. Finally, the *TSID*, which outperforms TSD and TSI on all three cases follows the same pattern as TSI and performs better on schools with higher sparseness.

To verify our results further we did some analysis of variance tests on our results. We consider as significant the factors or interactions with a level of significance (SL) lower as $\alpha = 0.05$. The diversification strategy is considered most influential factor with a SL <0.0001 and F-value of 194.79. The intra-moves are also considered influential with a SL of 0.0001 and F-value of 17.16; the combination of diversification and intra-moves is also significantly influential (SL = 0.0028 and F-value = 9.76)

Fig 6 illustrates cost function changes of the proposed algorithm over a run of 3000 iterations on *DE* test case. This figure shows that *TS* stops improvement very soon and does not improve after iteration 17. However other implementations improve overtime. *TSD* reaches its minimum at iteration 178, much faster than *TSI* at iteration 2322. However the combination of both yields best result at iteration 2664.

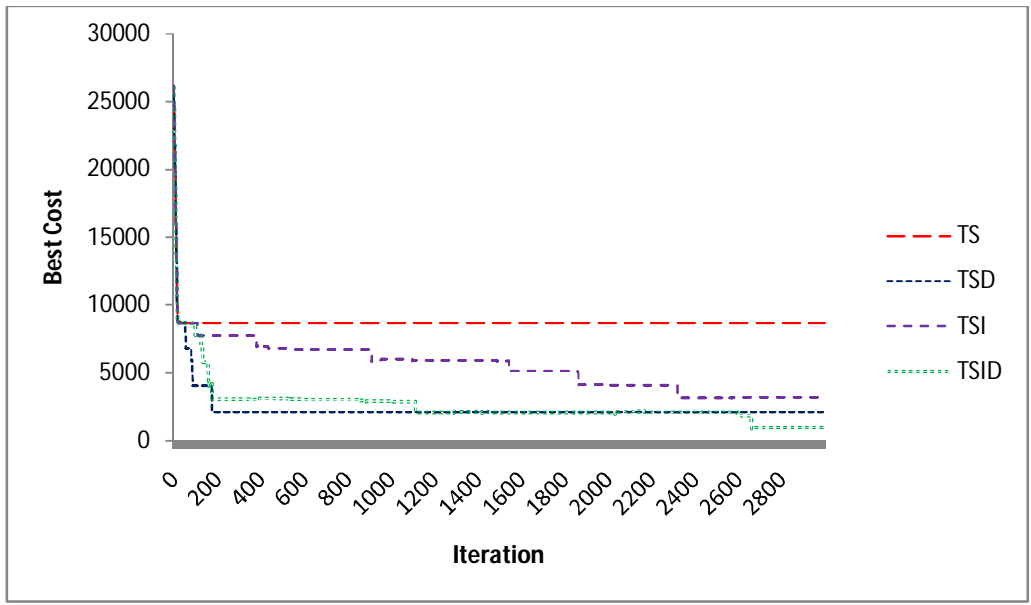

Fig 6: Best Cost over 3000 iterations

As described in section 3.3.2.1, intra moves controlled by IntraDepth parameter which updates its value dynamically. To verify this control mechanism we tested our algorithm considering fixed values for IntraDepth. Although it seems that higher intra-depth values could cause better results. Our experiment does not validate this claim. In DE school increasing intra-depth from 1 to 10 decreases the cost function by 7%; however the results get worse by more intra-depth; the cost function increase to 3070 at intra-depth of 20. It may be due to the fact that the algorithm searches in very sparse areas of search space and cannot take advantage of frequency based long term memories defined in section 3.2.5. It is also notable that higher intra-depth values have performance overheads since it takes much more time to explore the neighborhood. The best results achieved using adaptive intra-depth at 1050 and average value of intra-depth was 4.22 .

### 4.3 Analysis of Pre-Scheduling phase

As discussed in section 3.1. A graph based algorithm presented to find various types of blocks in lessons. Blocks, found by algorithm are validated by school's staff. Table 4 shows results of applying the algorithm on our dataset. The algorithm can find all blocks in Ta and *AL* dataset; however in *DE* school 72% of blocks hours are detected. To sum up, the algorithm can detect

32

92% of block hours on average. It is really a good performance for a feature which is really demanded by schools' staff.

Table 4: Results of applying pre-scheduling algorithm on detection of block sessions

| Test cases | Half-Switch | | Half-chain | |
| --- | --- | --- | --- | --- |
| | All | Detected | All | Detected |
| (TA) | 32 | 32 | 0 | - |
| (AL) | 4 | 4 | 2 | 2 |
| (DE) | 0 | - | 14 | 10 |

## 5 Conclusions and future work

This paper presented a new tabu search heuristic for high school timetabling problem. The algorithm is designed by collaboration of educational staff and considers requirements of real schools. The experiments shown that, this algorithm can build near optimal timetables acceptable by school's staff.

Contributions of the proposed algorithm include: using long term memory structures to guide diversification process of optimization phase which prevents the process to become entrenched in local optimum; In addition, adopting various types of moves and selecting them dynamically which makes the algorithm effective in different test cases. Moreover a graph based algorithm introduced to detect blocks in lessons as a pre-process for optimization. This algorithm was able to detect more that 90% of blocks on real datasets.

The proposed algorithm does not independent on data structures commonly used in timetabling research like matrix of teachers or classes over time period. However it generally assigns a set of events (lessons) to periods. Even, the diversification process and proposed moves do not dependent of those structures. Thus the algorithm has the potential to be used in other timetabling problems like university or nurse scheduling.

Although the algorithms performed well in all real data, using other tabu search components like intensification and strategic oscillation may improve it in terms of consistency and speed.

## Acknowledgements

We are thankful to the Hamedan University of Technology, Hamedan-Iran and research department of Hamedan's education organization for the support of this work.